\documentclass{article}

\usepackage[preprint]{neurips_data_2021}

\usepackage[utf8]{inputenc} 
\usepackage[T1]{fontenc}    
\usepackage{hyperref}       
\usepackage{url}            
\usepackage{booktabs}       
\usepackage{amsfonts}       
\usepackage{nicefrac}       
\usepackage{microtype}      
\usepackage{xcolor}         

\usepackage{epsfig}
\usepackage{graphicx}
\usepackage{amsmath}
\usepackage{amssymb}
\usepackage{subcaption}

\usepackage{biblatex}
\title{Multi-Dataset Benchmarks for Masked Identification using Contrastive Representation Learning}

\author{%
  Sachith Seneviratne \\ 
  University Of Melbourne\\
  Australia\\
  \texttt{sachith.seneviratne@unimelb.edu.au} \\
   \And
   Nuran Kasthuriaarachchi \\
   University of Moratuwa \\
   Sri Lanka\\
   \texttt{nuran.11@cse.mrt.ac.lk} \\
   \AND
   Sanka Rasnayaka \\
   School of Computing \\
   National University of Singapore \\
   \texttt{sanka@u.nus.edu}
}

\begin{document}

\maketitle

\begin{abstract}
The COVID-19 pandemic has drastically changed accepted norms globally. Within the past year, masks have been used as a public health response to limit the spread of the virus. This sudden change has rendered many face recognition based access control, authentication and surveillance systems ineffective. 
Official documents such as passports, driving license and national identity cards are enrolled with fully uncovered face images. However, in the current global situation, face matching systems should be able to match these reference images with masked face images. As an example, in an airport or security checkpoint it is safer to match the unmasked image of the identifying document to the masked person rather than asking them to remove the mask. We find that current facial recognition techniques are not robust to this form of occlusion.

To address this unique requirement presented due to the current circumstance, we propose a set of re-purposed datasets and a benchmark for researchers to use. We also propose a contrastive visual representation learning based pre-training workflow which is specialized to masked vs unmasked face matching. We ensure that our method learns robust features to differentiate people across varying data collection scenarios. We achieve this by training over many different datasets and validating our result by testing on various holdout datasets.
The specialized weights trained by our method outperform standard face recognition features for masked to unmasked face matching. We believe the provided synthetic mask generating code, our novel training approach and the trained weights from the masked face models will help in adopting existing face recognition systems to operate in the current global environment. We open-source all contributions for broader use by the research community.
\end{abstract}

Repo: \url{https://github.com/sachith500/ContrastiveFaceRepresentation}

\section{Introduction}

Facial recognition technology was generating impressive results prior to the COVID-19 pandemic. However, due to mask-based occlusions these methods now need to be investigated and adjusted to be robust to partial facial occlusion. A common scenario that occurs in this space is of unmasked vs masked identity matching. Often organizations will retain unmasked images of an individual appearing on various identity documents (passport, driver's license, staff identity) that need to be verified against a masked image. Traditional facial recognition methods contain feature representations that are reliant on seeing the whole face. In particular, the absence of some distinctive facial appendages in the masked image (lips, chin, moustache) is likely to lead to a false negative where an authentic user may be incorrectly categorized as an imposter. It is imperative that computer vision techniques are able to adapt to such scenarios. We find that resources for performing research in this domain are quite lacking and propose a set of benchmarks to remedy this situation.

Masked face recognition focuses on identifying people using their facial features while they are wearing masks. Masked facial recognition can be tackled across two use cases. First is to assume each user will enroll their face image while wearing a mask. This means matching is performed between two masked faces. The second use case is masked person recognition from a database of unmasked images. This use case is more receptive to using existing face databases such as passports or drivers' licenses. It has the broader advantage of not requiring an entire cohort of individuals to be re-registered within a facial database while wearing masks. Our work will focus on this scenario. 

We approach this problem of masked to unmasked matching with the objective of creating a replicable workflow that can be applied in the wild. To this end, we focus our analysis on evaluating on datasets unseen by the model during training. We re-purpose some existing and easily accessible facial databases with a synthetic masking technique in order to generate new datasets for this problem. We make our evaluation more robust by using several such databases and by performing additional evaluation on a new dataset collected explicitly for this problem. Our evaluation shows that our method outperforms existing facial recognition techniques even when finetuned on the same datasets. We use a workflow we believe is generalizable to benchmarking other problems in the occluded imagery domain, and therefore avoid using task-specific optimizations such as specialized loss functions from facial recognition and focus instead on improving performance by incorporating more datasets. By using a task-general workflow to optimize performance in a task-specific manner by using data as the instrument, we position this work for use as a general benchmarking technique in one-shot learning.





\section{Literature Review}

\subsection{Face mask recognition}

With the onset of COVID-19, the task of face mask attention has received considerable attention. Several studies have focused on classifying masked and unmasked faces achieving near perfect results over 99\% \cite{loey2021hybrid, snyder2021thor, lin2016masked, oumina2020control}. These works focus only on the presence of a mask but does not ensure the mask is worn properly. Batagelj et. al. \cite{batagelj2021correctly} introduce the Face-Mask Label Dataset (FMLD) to train models to see if a person is properly wearing a mask or not with over 97\% accuracy.

While the previously mentioned studies tackle the face mask recognition problem as a classification task, object detection based approaches utilize You Only Look Once (YOLO) approaches also report 94\% and 81\% average precision  \cite{cao2020maskhunter, loey2021fighting}. 

Our work is not focused on face mask detection, we focus on masked face recognition.

\begin{table*}[h]
    \centering
    \begin{tabular}{| p{0.16\linewidth} | p{0.4\linewidth} | p{0.20\linewidth} | p{0.14\linewidth} |} \hline
        Paper &  Approach & Evaluation Dataset & Result\\ \hline
        Hariri \cite{hariri2020efficient} & Occlusion removal approach and training on VGG16 architecture & RWMFD & 91.3\% (acc) \\ \hline
        
        Wang et. al \cite{wang2020masked} & Face-eye based multi granularity model & MFRD & 95\% (acc) \\ \hline
        
        Ejaz et. al \cite{ejaz2019implementation} & PCA features and distance metric & In house dataset (500 images) & 73.75\% (acc) \\ \hline
        
        Ding et. al \cite{ding2020masked} & Latent part detection model for discriminative parital feature learning & In house datasets MFV, MFI and Synth mask LFW & 97.9\%, 94.3\%, 95.7\% respectively \\ \hline
        
        Montero et. al \cite{montero2021boosting} & Multi-task ArcFace method & MFR2 & 99\% (acc) \\ \hline
        
        Mandal et. al \cite{mandal2021masked} & Transfer learning approach with the ResNet-50 architecture & RWMFD & 47.91\% (acc) \\ \hline
        
        Li et. al \cite{li2020look} & De-occlusion and knowledge transfer to create unmasked images for recognition & AR dataset & 95.4\% (acc) \\ \hline
        
    \end{tabular}
    \caption{Current masked face recognition research}
    \label{tab:currMFR}
    
\end{table*}

\subsection{Masked face recognition}
Due to the sudden widespread usage of face masks, existing face recognition systems have become less reliable. The effect of face masks on existing face recognition tasks was studied by Damer et. al. \cite{damer2020effect, damer2021masked}. Both these studies show quantitative evidence to show that current face recognition models drop in accuracy when the probe images are wearing masks. This highlights the need for specialized models which can handle masked faces without a drop in authentication accuracy.

Initial work on \textit{Occlusion robust Face Recognition (OFR)} \cite{oh2008occlusion, min2011improving, park2015partially, song2019occlusion} and \textit{Partial Face Recognition (PFR)} \cite{hu2013robust, weng2016robust, liao2012partial} has overlap with masked face recognition tasks. However, with the renewed importance of masked faces in the current global environment, there are several studies dedicated to \textit{Masked Face Recognition (MFR)}.

The main focus of existing MFR studies have been on developing new models which can do facial recognition for masked face datasets. Table \ref{tab:currMFR} gives a summary of current research on this area and the reported accuracy values. All the work summarized in Table \ref{tab:currMFR} focus on recognition where the reference and probe are both masked images. However, we focus on matching a unmasked reference image with a masked probe image.

Geng et. al \cite{geng2020masked} proposed a Generative Data Augmentation method to create synthesized data which is used to fine tune VGGFace2 model. In this work the authors evaluate the model on a scenario where the similarity between masked and unmasked faces are evaluated. They report a 86.5 F1 score for MFSR dataset. We will be focusing on a similar evaluation setup over many datasets.
Other research in this area includes masked face recognition using near IR images by Du et. al \cite{du2021towards}.

\subsection{Representation Learning}

Unsupervised representation learning has been explored extensively in computer vision due to the ability to learn from unlabelled images. This allows for a task-independent approach to representation building since unlabelled images are commonly available for most problems. Self-supervised representation learning is a type of unsupervised representation learning which performs unsupervised learning by creating a pretext task for the representation to be built in a supervised manner. Most self-supervised techniques vary in terms of the task used, including distortion\cite{Exemplar}, relative position prediction\cite{DoerschRelativePosition}, jig-saw puzzle solving\cite{NorooziJigsaw}, feature counting\cite{NorooziCounting}, coloring\cite{zhang2016colorful}. Current state of the art approaches in this area \cite{BYOL, chen2020simple, he2020momentum} use contrastive learning tasks to generate representations. We use MoCoV2 \cite{chen2020improved} in this work, which operates on the pre-text task of instance discrimination, as the basis for building our representations. In particular, we draw upon the idea of using a projection head during representation learning used by both SimCLR\cite{chen2020simple} and MoCoV2, followed by discarding it during evaluation and extend this idea to replicating this workflow at inference time.


\section{Methodology}

\begin{figure*}
\centering
\begin{subfigure}{.25\textwidth}
  \centering
  \includegraphics[height=.7\linewidth]{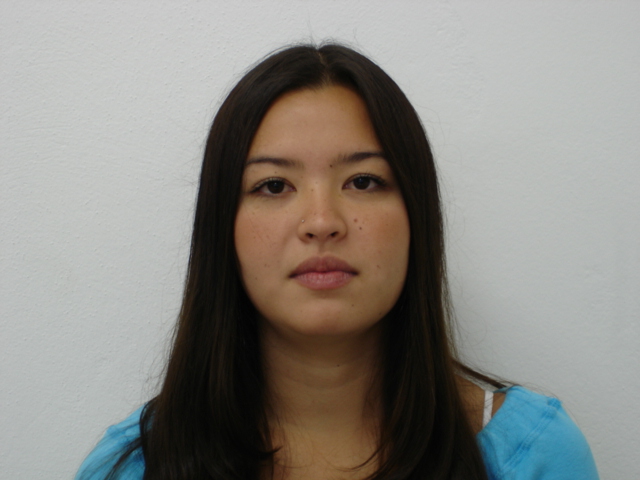}
  \caption{Original image}
  \label{fig:sfig1}
\end{subfigure}%
\begin{subfigure}{.25\textwidth}
  \centering
  \includegraphics[height=.7\linewidth]{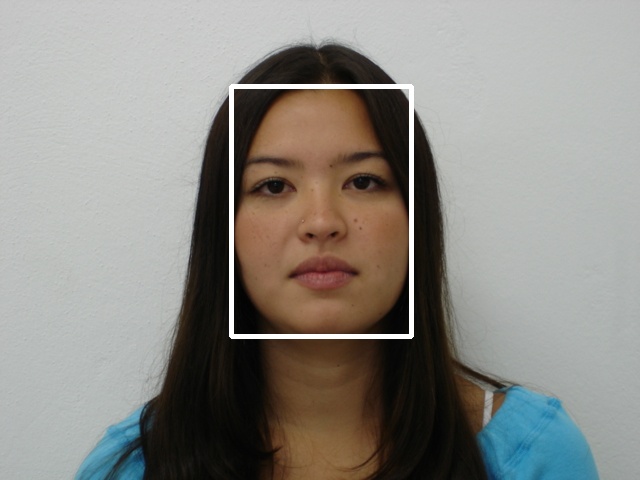}
  \caption{Face localization}
  \label{fig:sfig2}
\end{subfigure}
\begin{subfigure}{.2\textwidth}
  \centering
  \includegraphics[height=\linewidth]{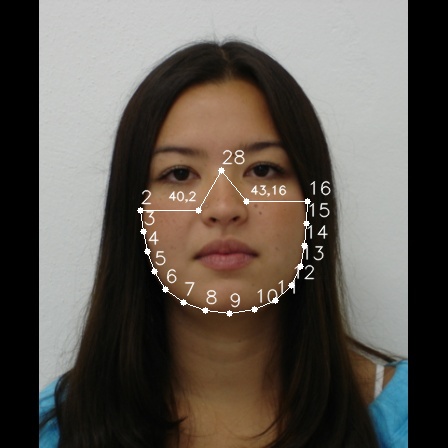}
  \caption{Key point detection}
  \label{fig:sfig2}
\end{subfigure}
\begin{subfigure}{.2\textwidth}
  \centering
  \includegraphics[height=\linewidth]{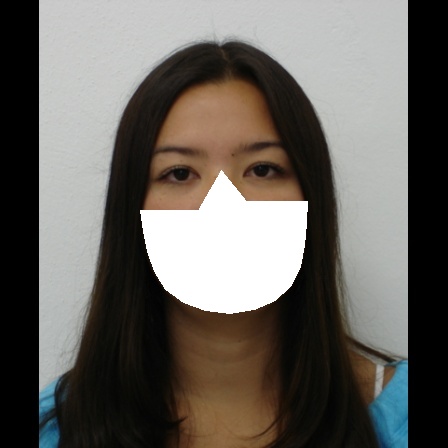}
  \caption{Digital mask added}
  \label{fig:sfig2}
\end{subfigure}
\caption{Synthetic mask creation pipeline (example for a Georgia Tech Face image)}
\label{fig:digimask}
\end{figure*}



\subsection{Datasets}

We use two approaches to create masked faces for training and testing. (1) Use existing large scale face datasets by adding a digital mask synthetically. (2) Collect a small scale dataset of masked and unmasked images from volunteers for validation.

\subsubsection{Creating Synthetic Masks}
We follow the process proposed by Ngan et. al \cite{ngan2020ongoing} to draw a digital mask on top a facial image. First we detect the frontal face bounding box using the face detector from \cite{dlib09}. After cropping the face we use facial key point predictor from \cite{dlib09} using 68 facial key points. A synthetic mask shape is created by generating the convex hull by combining selected key points. The intermediate steps of this process are depicted in Fig. \ref{fig:digimask}. All steps are reproducible for any data using scripts we open-source\footnote{https://github.com/sachith500/ContrastiveFaceRepresentation}. Masking is verified by performing landmark detection on the resultant masked images. Masked images for which a facial bounding box is not detected are discarded. This is so that face detection workflows can still work with such images.

\subsubsection{Collecting Validation Dataset}
\begin{figure}[h]
\centering
\begin{subfigure}{.25\textwidth}
  \centering
  \includegraphics[width=.8\linewidth]{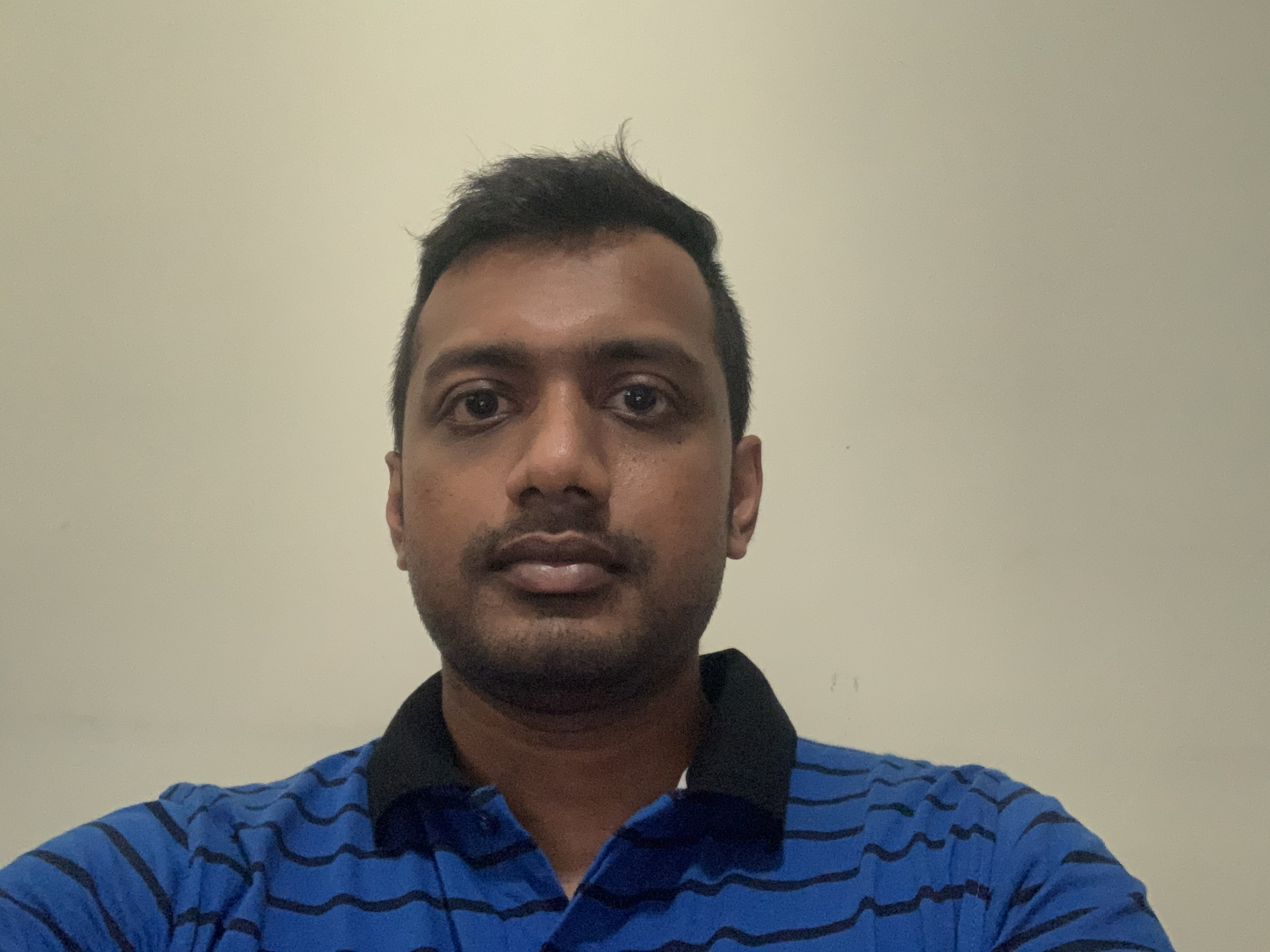}
  \caption{Non masked}
  \label{fig:sfig1}
\end{subfigure}%
\begin{subfigure}{.25\textwidth}
  \centering
  \includegraphics[width=.8\linewidth]{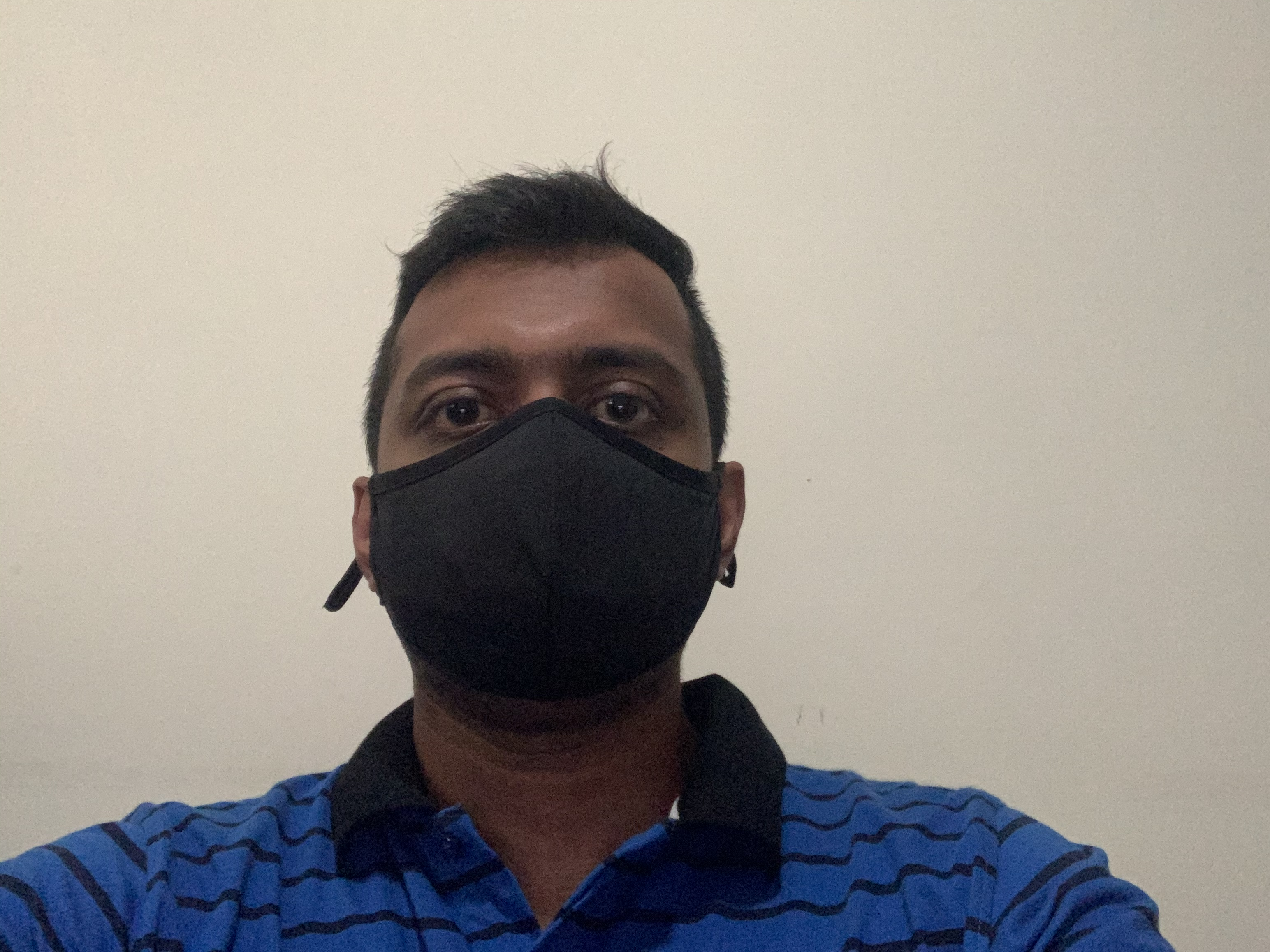}
  \caption{Masked}
  \label{fig:sfig2}
\end{subfigure}
\caption{Example image pair from in house dataset}
\label{fig:ourdataset}
\end{figure}

Since the training data was created by adding a digital mask over an un-masked face, we collect a real dataset with masked and unmasked images for each identity. This collection is done on a voluntary basis where the participants are shown an example pair of images and asked to capture themselves using the front camera of their mobile device.
This created a challenging dataset of varying lighting conditions, indoor/outdoor environments, different mask types and different camera qualities. Therefore, this validation dataset gives a good indication of how robust and genaralizable our models are. An example image pair is shown in Fig. \ref{fig:ourdataset}.

\begin{table}[]
    \centering
    \begin{tabular}{|l|l|l|} \hline
        Dataset  &  Unmasked Identities/ Images & Masked Identities/ Images\\ \hline
        CelebA \cite{liu2015faceattributes} & 10177/202,599 & 10174/197,499\\ \hline
        FEI Face \cite{THOMAZ2010902} & 200/1,177 & 200/1,177 \\ \hline
        Georgia Tech \cite{georgiatech} & 50/750 & 50/750\\ \hline
        SoF \cite{afifi2017afif4} & 93/1,443 & 90/1,393\\ \hline
        YouTube Faces \cite{wolf2011face} & 1595/20,252 & 1589/19,960\\ \hline
        LFW \cite{LFWTech} & 5749/13,167 & 5718/13,138\\ \hline
        In house Dataset & 41/41 & 41/40\\ \hline
    \end{tabular}
    \caption{Summary of datasets used for training and evaluation}
    \label{tab:datasets}
\end{table}

Table \ref{tab:datasets} gives a summary of the datasets which were created and collected in this study. We use CelebA, LFW, YouTube Faces and SoF with a train and test split for training the model and testing. We keep FEI Face, Georgia Tech and In house dataset as holdout sets to validate our models' generalisability.

\subsection{Model Architecture}
\label{Methodology:Architecture}
\begin{figure}
\centering
\includegraphics[width=0.7\linewidth]{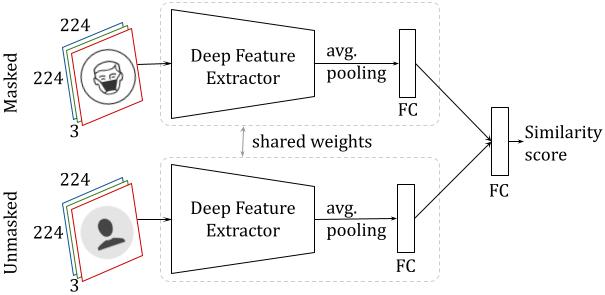}
\caption{Siamese Network Architecture}
\label{fig:siamese}
\end{figure}

We use a Siamese network with shared weights as the basis of all training workflows. The embedding outputs from standard model architectures (ResNet,VGG,MobileNet) etc, are used to compute the distance between an image pair (masked vs non-masked). This is then fed into an intermediate fully connected layer with sigmoid activation, which is connected to a final output with linear activation. Training is done using binary cross-entropy and similarity is measured at inference time using one of three methods. Where distance between two vectors is used as a measure of dis-similarity, we convert to similarity as $\frac{1}{1 + Distance}$ which is in [0,1] for Distance in [0,$\infty$).

\begin{itemize}
    \item Similarity output at the output level - output is passed through a sigmoid at inference time only to scale to [0,1]
    \item Similarity as a function of L2 distance at the intermediate fully connected level (generally 512 nodes with sigmoid activation)
    \item Similarity as a function of L2 distance at the bottleneck/embedding layer of the backbone architecture. (for example a vector of 2048 dimensions in ResNet50).
    
    Figure~\ref{fig:farfarr2} in Section~\ref{Results:ModelOptimization} has an example characteristic response curve for these options.
    
\end{itemize}

\subsection{Training Setup}
As we use a Siamese network based approach for training our feature extractor,  we create pairs of images for training. Each pair corresponds to an unmasked reference and a masked probe image. The network outputs a similarity in [0,1] with 0 indicating imposter and 1 indicating authentic match. Since absolute difference is taken between embeddings from a shared weight siamese network, the ordering of masked/unmasked images as reference and probe has no effect on the final similarity scores. Figure~\ref{fig:siamese} contains a high level overview of the architecture.


\subsection{Training Workflow}

The training workflow primarily proceeded as follows:

\begin{itemize}
    \item Use a pretrained representation to build a model on a single dataset.
    \item Finetune the built model on multiple datasets to generalize the feature embedding.
    \item Further finetune based on identifying hard negative pairs during training.
\end{itemize}

Training was carried out on image pairs, drawn at random from the training set of identities. The shared weight Siamese formulation mentioned above was used, with an additional linear layer connected to a sigmoid activation function operating on the L2 distance between the embeddings of each image. Binary cross entropy was used as the training loss.
Pretrained representations were obtained through several means, many were found from facial recognition/detection tasks from existing literature, custom representations using unsupervised learning were also generated and tested. Finetuning was carried out by freezing parameter updates for part of the network and using validation results as an indication of embedding improvement. A custom representation was generated using MoCoV2 with the following training parameters (recommended in \cite{chen2020improved}): learning rate 0.015, batch-size 128 with MoCo softmax temperature 0.2, while using the projection head and augmentation workflow introduced in SimCLR. Pre-training was carried out on a 4 GPU node on Spartan\cite{lafayette2016spartan}  for 860 epochs on the CelebA(Masked and Unmasked) dataset. The resultant representation was finetuned end-to-end on 4 datasets (CelebA, LFW, SoF, youtube - masked and unmasked) combined for another 25 epochs by continuing the pretraining process. This exposes the representation to easier negatives (cross-dataset) and provides more data variety for the pretraining process. There were several approaches considered in this regard: for example, given in this context that most of the facial features being matched are outside of the masked region, it is possible to argue that using only masked images would be sufficient for representation learning. However, since the reference image is unmasked and the level of occlusion of facial features depends on many factors such as type of mask and the assumption that it is worn properly, we decide to include both masked and unmasked images in learning features. The primary impact this has on the pretraining workflow is that the model gains contextual knowledge regarding both masked and unmasked images and learns features to be able to distinguish between the two types. This stems from the fact that the pretraining workflow is focused around instance discrimination. Additionally, this creates a representation which should theoretically be useful for extending to other tasks. We share all representations (initial and final) used in this paper for further research in this area.

\textbf{Validation} was done using a precision metric. From the validation set of identities, a single identity(unmasked) is chosen as the reference and a masked image is drawn from the same identity forming an "authentic pair". 19 identities are drawn uniformly at random with replacement from the set of available identities excluding the reference identity. From these "imposter" identities 1 image is drawn per identity uniformly at random forming 1 authentic pair and 19 imposter pairs following the workflow in \cite{lake2011one}. Evaluation on 20 such pairs counts as one validation step. 400 such steps are conducted at the end of each training iteration. Precision over the iteration is counted as the percentage of steps where the authentic image pair has the highest similarity (out of 20 possible pairs). Training iterations which produce a checkpoint with at least 90\% validation precision were chosen for further evaluation on holdout datasets. Note that the expected precision from a random prediction would be 5\% in this case ($\frac{1}{20}$). The similarity in this experiment is always inferred from the final linear layer. Due to the quadratic scaling (${n \choose 2}$ where n is number of identities) of imposter pairs, this allows us to use more identities in training with fewer used for validation.

\subsection{Evaluation}

\subsubsection{Evaluation Metrics}
Our models output a similarity score for a given masked and unmasked image pair. Therefore, the decision outcome of the system depends on a threshold value.

\begin{equation}
            \text{if similarity(reference, probe)} >= \text{threshold}: \normalsize\raggedleft 
            \text{Accept as legitimate user}
\end{equation}

With this setup, there is a trade off between false accepts and false rejects as we alter the threshold value. Therefore, the evaluations are done by measuring following metrics. 

\begin{itemize}
    \item \textbf{False Acceptance Rate (FAR)/False Reject Rate (FRR):} We analyze the trade off between FAR and FRR by plotting these two with respect to different threshold values.
    \item \textbf{Equal Error Rate (EER):} We calculate the error rate where FAR and FRR are equal as a quantitative measure of accuracy.
    \item \textbf{FRR100:} The lowest FRR for a False Acceptance Rate (FAR) $<1.0\%$
\end{itemize}

\subsection{Holdout Testing}

We reserve several datasets for the purpose of holdout testing. We generate an equal number of authentic and imposter pairs randomly and fix them for evaluation. These lists are released as part of our benchmark for one to one comparisons in future studies.

\subsection{Benchmark 1: Representation evaluation}

We performed an experiment in order to select a model backbone architecture as well as a training workflow. We evaluated several different models using the training approaches mentioned previously, and built with different pretrained representations to use as starting points for training. Similarity was measured directly at the final sigmoid layer of the output. Our objective was to identify the smallest model that was capable of generalizing good results over multiple datasets while also exploring the merits of training a new baseline representation useful for masked classification tasks. We select the best model checkpoint based on validation performance from each training workflow and compare them by performing inference on a comprehensive set of holdout datasets. We propose this as a benchmark for learning effective masked facial representations from a single dataset using masked/unmasked images, while evaluating on multiple holdout datasets. This benchmark captures the capacity of a particular model training process to generate data-set independent representations suitable for use in the wild. CelebA is uniquely suited for training purposes as there is more variation present in terms of within-identity age, hair style, pose and emotion variation. This is important in unmasked-masked identification as methods need to be robust to changes in all these factors (reference unmasked images are often used for a while before being recaptured).

\subsection{Benchmark 2: Task evaluation}

In this benchmark we explore the capacity of a training workflow to utilize multiple datasets in order to learn a feature representation with a focus on following a general workflow that can be useful in other tasks involving one-shot learning.

\subsubsection{Stage 1: Multi-dataset training}

The selected workflow from the previous experiment was used to finetune the pretrained model using 4 datasets: CelebA, LFW, sof\_original and youtube. First, we freeze 50\% of the ResNet50 and use Stochastic Gradient Descent with a learning rate of 1.0 to learn an overall strong representation. Due to training on 4 datasets at once and freezing a large part of the network, overfitting is avoided. The high learning rate allows the training prcess to explore the hypothesis space quickly and possibly cause gradient descent to bounce out of local minima, while our validation precision metric serves to identify checkpoints which have learnt a strong feature embedding. At this stage, we filter checkpoints with precision 90\% or higher for further evaluation. Imposter pairs are drawn from within the same dataset (to minimize the model's focus on image background features). The dataset to draw a particular pair from is drawn as a categorical variable with selection probability directly proportional to the size of each dataset. All datasets not used for training and validation were used as holdout datasets.

We isolate two such promising checkpoints for further finetuning (CP1 and CP2) as the two checkpoints with highest precision. These models are further finetuned with a low learning rate across a grid of parameter settings to perform further improvements to prediction performance. The three such best performing versions are used to create a simple averaging ensemble, which averages the similarity of each individual model at inference time. We additionally incorporate training on harder imposter pairs during training. For this we first draw an identity to serve as the reference, and then draw a number of imposter images from the same dataset. Inference is carried out on these images to identify the pair that is hardest for the model to classify as an imposter pair (i.e. the pair with the highest similarity), and then training is carried out on this pair. As this happens during the data-loading within the training pipeline, it considerably slows down the training process but provides a way for the model to learn on imposter pairs it is currently most likely to misclassify as authentic. 

For combining datasets for training, we use two main strategies for drawing training pairs from the datasets. \textbf{Uniform} sampling refers to drawing a dataset with probability proportional to the dataset size (thus representing one large dataset as opposed to 4 individual datasets, but without drawing imposter pairs from different datasets). \textbf{Stratified} sampling refers to drawing the dataset to sample from uniformly at random, thereby ensuring that each dataset is represented equally. We primarily use uniform sampling for model exploration and stratified sampling for model finetuning.

\begin{table*}[h]
    \centering
    \begin{tabular}{|l|l|l|l|l|l|l|l|} \hline
        Model & Base & Iterations & Batch Size &  LR & Frozen\% & hard sample size & draw strategy\\ \hline
        CP1 & -  & 695000 & 128 & 1.0 & 50 & - &  uniform\\ \hline
        CP2  &  -  & 885000 & 128 & 1.0 & 50 & - &  uniform\\ \hline
        FT1  &  CP1 & 11001 & 32 & 0.001 & 90 & hardest of 16  &  stratified\\ \hline
        FT2  &  CP1 & 11251 & 32 & 0.01 & 80 & hardest of 32  &  stratified\\ \hline
        FT3  &  CP2 & 14501 & 32 & 0.01 & 50 & hardest of 10 & stratified\\ \hline
    \end{tabular}
    \caption{Experiment 2 parameter settings}
    \label{tab:pairs}
\end{table*}

We combine \textbf{FT1}, \textbf{FT2} and \textbf{FT3} as a simple ensemble to derive the final benchmark. Inference is performed on each individual image pair by each \textbf{FT} model and the similarities are derived as $\frac{1}{1+Distance}$ with euclidean distance taken at the 2048 bottleneck layer of each model. The resultant similarities are averaged to create the final similarity result.
\section{Results}

\subsection{Benchmark 1: Representation evaluation and Model selection}
\label{Results:ModelSelection}

The results in Table~\ref{tab:benchmark1} indicate training results under equal levels of training. ImageNet and VGGFace2 are finetuned end to end but the contrastive masked representation trained on CelebA is superior.
\begin{table*}[h]
    \centering
    \begin{tabular}{|l|l|l|l|l|l|l|} \hline
        &\multicolumn{2}{|l|}{ImageNet}&\multicolumn{2}{|l|}{VGGFace2}& Proposed \\ \hline
        Dataset & VGG19 & MobileNet & SENET & VGG16 & ResNet50 \\ \hline
        fei\_face\_original & 0.363 & 0.356     & 0.49  & 0.304 & \textbf{0.031}    \\ \hline
georgia\_tech       & 0.323 & 0.416     & 0.483 & 0.431 & \textbf{0.097}    \\ \hline
sof\_original       & 0.476 & 0.389     & 0.415 & 0.365 & \textbf{0.169}    \\ \hline
fei\_face\_frontal  & 0.357 & 0.171     & 0.424 & 0.143 & \textbf{0}        \\ \hline
youtube\_faces      & 0.424 & 0.394     & 0.468 & 0.385 & \textbf{0.115}    \\ \hline
lfw                 & 0.361 & 0.449     & 0.469 & 0.372 & \textbf{0.142}    \\ \hline
in\_house\_dataset  & 0.288 & 0.244     & 0.425 & 0.288 & \textbf{0.038}    \\ \hline
    \end{tabular}
    \caption{Benchmark 1: Equal Error Rates of different models across various datasets initialized with different pretrained representations, trained on CelebA for 164k steps.}
    \label{tab:benchmark1}
\end{table*}

\subsection{Benchmark 2: Multi-dataset training benchmark}
\label{Results:ModelOptimization}

\begin{figure}[h]
\centering
\begin{subfigure}{.5\textwidth}
  \centering
  \includegraphics[width=\linewidth]{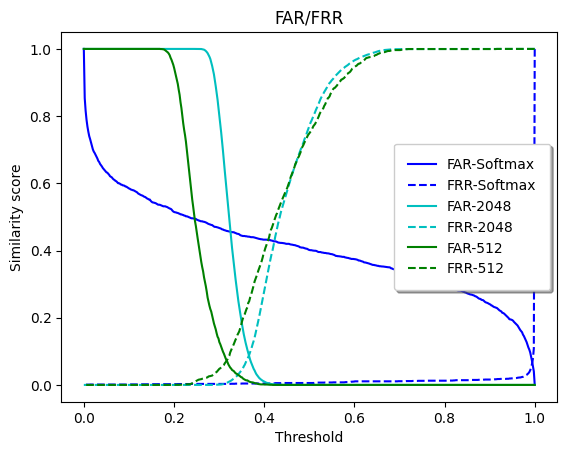}
  \caption{FAR/FRR curve, for all datasets}
  \label{fig:sfig1}
\end{subfigure}%
\begin{subfigure}{.5\textwidth}
  \centering
  \includegraphics[width=\linewidth]{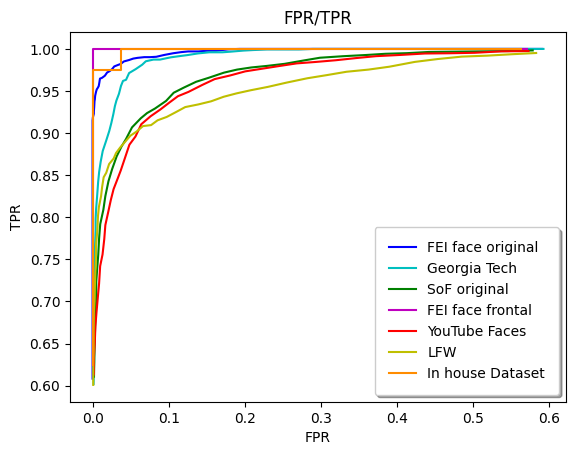}
  \caption{FPR/TPR curve, for all datasets}
  \label{fig:sfig2}
\end{subfigure}
\caption{(a) justifies use of 2048 layer for inference as it provides better intruder detection. (b) provides an indication of task difficulty on each dataset (Lower AuC = more difficult)}
\label{fig:farfarr2}
\end{figure}

The results of using the different inference strategies from Section~\ref{Methodology:Architecture} can be seen in Figure~\ref{fig:farfarr2}. We find that deriving similarity at the 2048 or 512 levels results in a better spread of the values compared to applying a softmax function on the final output. Comparing between 512 and 2048 levels we see that while the equal error rate is similar there is a drop in the FRR rate of 2048 features compared to 512. Therefore, we use the 2048 bottleneck features for future experiments.

\begin{table*}[h]
    \centering
    \begin{tabular}{|l|l|l|l|l|l|l|l|} \hline
        Dataset & Exp1 & CP1 & CP2 & FT1 & FT2 & FT3 & Ensemble\\ \hline
fei\_face\_original & 0.073  & 0.016 & 0.015 & 0.01  & 0.016 & 0.011 & \textbf{0.009} \\ \hline
georgia\_tech       & 0.207 & \textbf{0.041} & 0.055 & 0.06  & 0.059 & 0.058 & 0.048 \\ \hline
\textit{sof\_original}       & 0.187 & 0.073 & 0.071 & \textbf{0.058} & 0.069 & 0.067 & 0.061 \\ \hline
fei\_face\_frontal  & 0 & 0     & 0     & 0     & 0     & 0     & 0     \\ \hline
\textit{youtube\_faces}      & 0.156 & 0.053 & 0.051 & 0.042 & 0.056 & 0.046 & \textbf{0.041} \\ \hline
\textit{lfw}                 & 0.219 & 0.101 & 0.09  & 0.091 & 0.11  & 0.093 & \textbf{0.084} \\ \hline
in\_house\_dataset  & 0.038 & 0.031 & \textbf{0.013} & \textbf{0.013} & \textbf{0.013} & 0.019 & \textbf{0.013}\\ \hline
    \end{tabular}
    \caption{Benchmark 2: EER results. Exp1 refers to the proposed model from experiment 1 trained for 164k steps with inference done at 2048-level. Datasets in Italic are used for training in all models except Exp1. Table~\ref{tab:pairs} has the parameters for each model.}
    \label{tab:results}
\end{table*}




The relative difficulty of different datasets can be visualized using Figure~\ref{fig:farfarr2}. The overall benchmark consisting of single dataset based and multi-dataset based training is presented in Table~\ref{tab:resultsoverall}. We include the FRR100 metric which is more important for intruder detection in practical situations.
\begin{table*}[h]
    \centering
    \begin{tabular}{|l|l|l|l|l|} \hline
        &\multicolumn{2}{|l|}{Exp1 (Celeb Only)}&\multicolumn{2}{|l|}{Ensemble (4 datasets)}\\ \hline
        Dataset & EER & FRR100  & EER & FRR100 \\ \hline
fei\_face\_original & 0.089984 & 0.638723  & 0.008984 & 0.015968 \\ \hline
georgia\_tech       & 0.142884 & 0.93014   & 0.047976 & 0.245509 \\ \hline
sof\_original       & 0.195122 & 0.762745  & 0.061094 & 0.178431 \\ \hline
fei\_face\_frontal  & 0.071429 & 0.2       & 0        & 0        \\ \hline
youtube\_faces      & 0.142902 & 0.904     & 0.040948 & 0.208    \\ \hline
lfw                 & 0.17788  & 0.976048  & 0.08387  & 0.229541 \\ \hline
in\_house\_dataset  & 0.075    & 0.15      & 0.0125   & 0.025    \\ \hline
    \end{tabular}
    \caption{Overall benchmark: results of Exp1(trained on CelebA for 1015k steps) and Ensemble (trained on 4 datasets) on the synthetic unmasked-masked datasets generated}
    \label{tab:resultsoverall}
\end{table*}
\section{Conclusions and Future Work}

In this work we have presented techniques for synthetic data generation and analysis for masked facial recognition. We use a general framework for benchmarking that builds upon contrastive representation learning without specializing any methodology for facial analysis (beyond using facial data). We present 2 benchmarks on masked recognition across multiple synthetic datasets in an easily reproducible manner to facilitate further research in this area.
Our experiments show that using custom masking is an efficient way of creating datasets for training mask related models. We find that existing pretrained facial models appear to not be disentangled to the level where retraining them for use with masked images is straightforward. We show that it is better to use contrastive representation learning to build an initial representation and then adapt it to learn on the required facial task. We hypothesize that this is because existing identity/facial features use combinations of facial appendages - thus when several of them (nose, mouth, cheeks, lips, chin) are removed, the representations are unable to recover easily using simple fine-tuning. By training a fresh representation, it is possible to circumvent this issue of disentangling an existing representation by instead training the neural network to focus on what is present in the images rather than learn to ignore what is absent.

While previous work focuses solely on masked recognition, unmasked-masked recognition has several important use cases as standard identification protocols involve using unmasked base images. For example, all passports are taken with the face uncovered and with all facial features visible and there is additional information that can be derived from a fully visible face which can be useful for recognition with partially,incorrectly or fully masked faces. The datasets we synthesize and release provide the first publicly available and usable source of data conducive to this problem.


{\small
\printbibliography
}

\end{document}